\documentclass[letterpaper, 10 pt, conference]{ieeeconf}  % Comment this line out if you need a4paper

\IEEEoverridecommandlockouts                              % This command is only needed if 
\usepackage{graphics} % for pdf, bitmapped graphics files
\usepackage{amsmath, bm} % assumes amsmath package installed
\usepackage{amssymb}  % assumes amsmath package installed
\usepackage{color}
\usepackage{textcomp}
\usepackage{gensymb}
\usepackage{graphicx}

\usepackage{multicol,tabularx,capt-of}
\usepackage{multirow}

\usepackage[ruled,vlined]{algorithm2e}

\newcommand{\tablelidar}{
\begin{table}[tbp]
\caption{Specifications of Different LiDARs Used in Calculation}
\vspace{-1em}
\centering
\scriptsize
\begin{tabular}{|c|c|c|c|c|}
\hline
& \begin{tabular}{@{}c@{}} \# of \\Channels \end{tabular}   &  \begin{tabular}{@{}c@{}}Vertical \\Resolution \end{tabular}  & \begin{tabular}{@{}c@{}}Horizontal \\Resolution \end{tabular}    & \begin{tabular}{@{}c@{}}Vertical \\FOV \end{tabular}                       \\ \hline
Pandar64  & 64             &  $0.167^{\circ} \sim 6^{\circ}$        & $0.2^{\circ}$         & $[-25.0^{\circ}, 15.0^{\circ}]$     \\ \hline
HDL-64E  & 64             & $0.427^{\circ}$          & $0.167^{\circ}$         & $[-24.9^{\circ},2.0^{\circ}]$ \\ \hline
Pandar40 & 40             & $0.33^{\circ} \sim 6^{\circ}$        & $0.2^{\circ}$         & $[-25.0^{\circ}, 15.0^{\circ}]$     \\ \hline

\end{tabular}

\vspace{-2em}
\label{table:spec}
\end{table}
}

\newcommand{\tableresult}{
\begin{table}[tbp]
\caption{Optimal Sensor Placement and Corresponding Perception Entropy for Different Sensor Specifications }
\vspace{-0.5em}
\centering
\scriptsize
\begin{tabular}{|c|c|c|c|c|c|c|}
\hline
    \begin{tabular}{@{}c@{}} \# \\ \end{tabular} &\begin{tabular}{@{}c@{}}Sensor \\Spec\end{tabular}   & $t_x$(m)  & $t_y$(m) & $t_z$(m)                              & \textit{pitch}($^\circ$)            &  \begin{tabular}{@{}c@{}}perception \\entropy\end{tabular} \\ \hline  \hline
    
    \multirow{2}{*}{1} &  Pandar64   & $-0.43$       & $0$      &    $1.80$  &  $0$     &  $1.6429$  \\ \cline{2-7} 
         &HDL-64E & $-0.37$  & $0$    & $2.12$      &  $0$        &  $2.1212$       \\ \hline \hline

     \multirow{2}{*}{2} &  60\degree HFOV   & $1.00$       & $0$      &    $1.19$  &  $1.0$     &  $2.0055$   \\ \cline{2-7} 
 & 120\degree HFOV & $0.84$  & $0$    &  $1. 27$    &  $6.1$        &  $2.0237$       \\ \hline \hline
 
      \multirow{4}{*}{3} &  Pandar40a   &$3.03$       & $-1.01$      &    $1.16$  &  $-0.3$ &    \multirow{2}{*}{$1.7119$}   \\ \cline{2-6} 
 &Pandar40b & $-3.03$  & $1.01$    &  $0.85$     &  $3.2$ &          \\ \cline{2-7}   
 
       &  Pandar40a   & $3.03$       & $1.01$      &    $1.05$  &  $0.3$     &     \multirow{2}{*}{$1.6864$}   \\ \cline{2-6} 
 &Pandar40b & $3.03$  & $-1.01$    &  $1.03$     &  $0.1$        &       \\ \hline \hline
 
   \multirow{4}{*}{4} &  Pandar40a   & $3.03$       & $1.01$      &    $0.91$  &  $1.2$ &    \multirow{4}{*}{$0.8965$}   \\ \cline{2-6} 
 &Pandar40b & $3.03$  & $-1.01$    &  $0.84$     &  $3.0$ &          \\ \cline{2-6}   
 
       & 30\degree HFOV    & $2.90$       & $0$      & $1.66$  &  $1.4$   &   \\ \cline{2-6} 
 &60\degree HFOV & $-2.90$  & $0$    &  $1.52$     &  $7.2$        &       \\ \hline 
  
\end{tabular}

\vspace{-2em}
\label{table:result}
\end{table}
}

\newcommand{\alg}{
\begin{algorithm}[tbp]
\SetAlgoLined
\KwIn{ initial placement $\mathbf{q_0}$, initial neighborhood $N$,}
 neighborhood lower bound $N_0$, decay factor $\mathrm{k}$ \\
\KwOut{optimal placement $\hat{\mathbf{q}} $ with entropy $H_{min}$} 
 
 $\hat{\mathbf{q}} = \mathbf{q_0}$, $H_{min} = H(\mathbf{S}|M,\mathbf{q_0})$\\
  uniform sample perception space as set $S$\\
 \While{$N > N_0$}{ 
 random sample sensor placement around $\mathbf{q}$ within neighborhood $N$ as set $Q$\\
\For{$\mathbf{q}$ in $Q$}{
  $H_{total},p_{total} = 0$\\
  \For{$\mathbf{s}$ in $S$}{
  $\mathbf{m} = f(\mathbf{s},\mathbf{q})$\\
  apply early fusion on $\mathbf{m}$ to obtain $m_{fused}$ and estimate  $AP$\\
  calculate $\bm{\sigma}$ based on estimated $AP$\\
  apply late fusion on $\bm{\sigma}$ to obtain $\sigma_{fused}$\\
  $H(\mathbf{X}|\mathbf{m},\mathbf{q}) = 2\mathrm{ln}(\sigma_{fused}) +1+\mathrm{ln}(2\pi)$\\
  $H_{total} = H_{total} + p_\mathbf{S}(\mathbf{s})H(\mathbf{S}|\mathbf{m},\mathbf{q})$\\
  $p_{total} = p_{total} + p_\mathbf{S}(\mathbf{s})$
  } 
  \If{$ H_{total} / p_{total} >H_{min}$}{
    $\hat{\mathbf{q}} = \mathbf{q}$, $H_{min} =  H_{total} / p_{total}$\\ 
   }  
   } 
    $ N = \mathrm{k} \cdot N$  
 }
 \caption{Optimizing Installation Placement Given Sensor Selection}
 \label{alg:1}
\end{algorithm}
}

\newcommand{\figcurve}{
\begin{figure}[btp]
\centering
\includegraphics[width=0.95\linewidth]{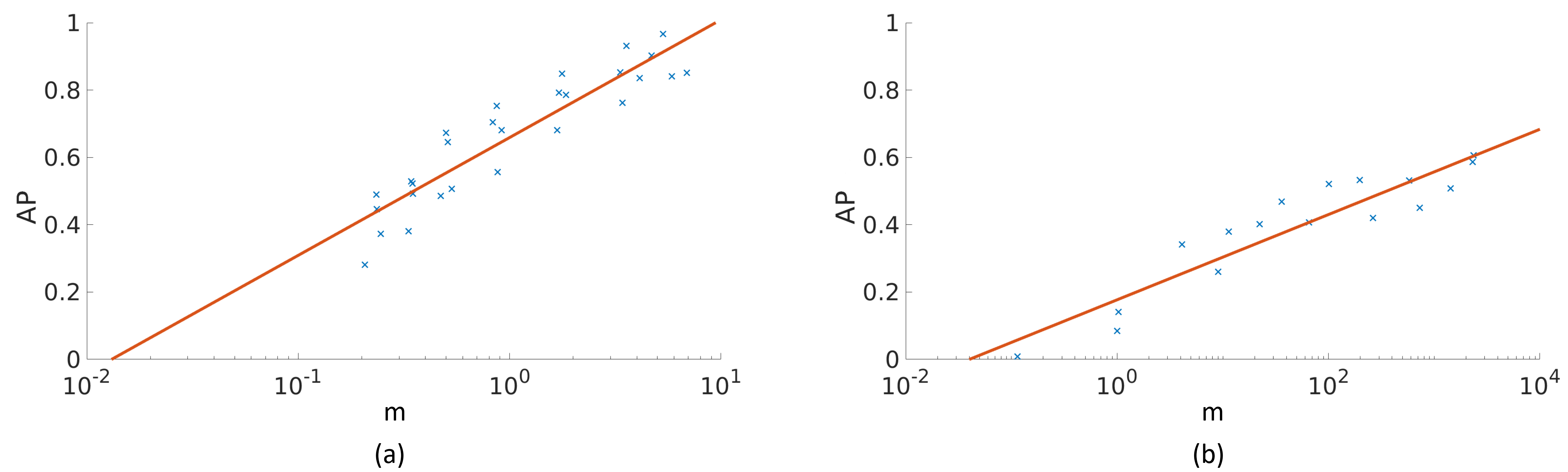}
\vspace{-1em}
\caption{Curve fitting for AP between LiDAR measurement (a) and camera measurement (b). }
\vspace{-1.5em}
\label{fig:curve}
\end{figure}
}

\newcommand{\figspace}{
\begin{figure}[btp]
\centering
\includegraphics[width=0.85\linewidth]{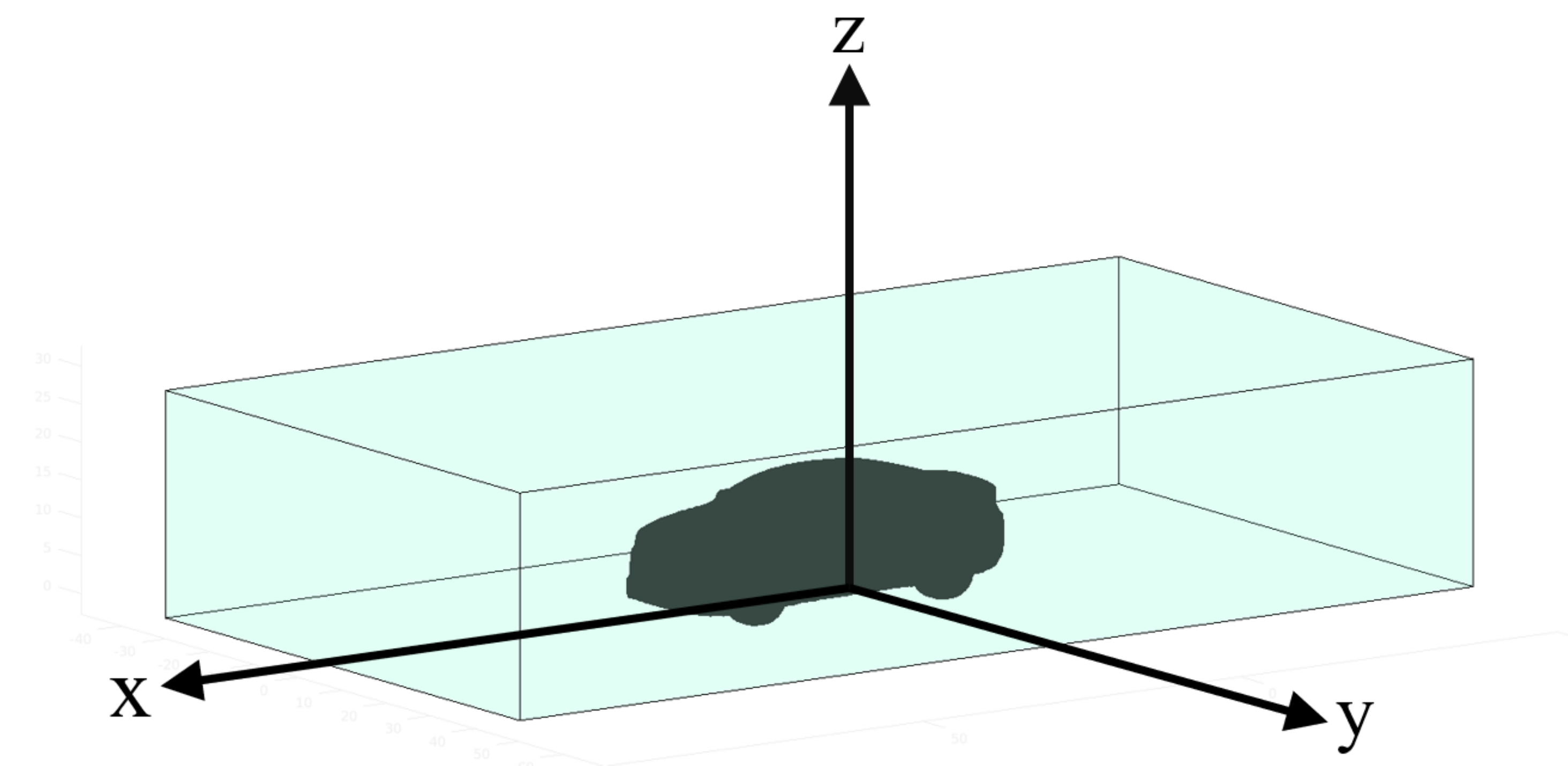}
\vspace{-1em}
\caption{The perception space.}
\vspace{-1.5em}
\label{fig:space}
\end{figure}
}

\newcommand{\figvoxelize}{
\begin{figure}[btp]
\centering
\includegraphics[width=0.85\linewidth]{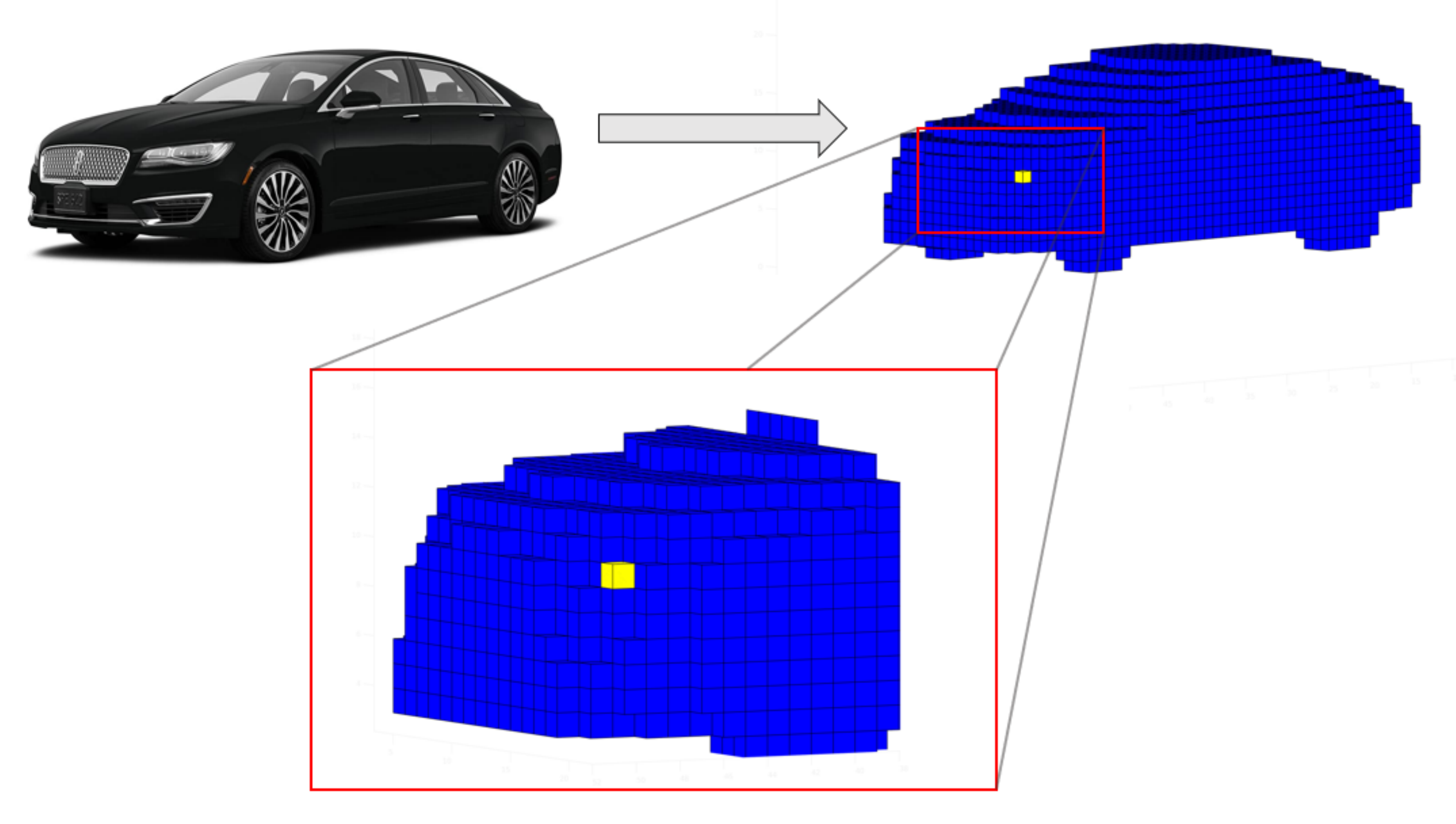}
\vspace{-1em}
\caption{The car is voxelized into small cubes and its perception potential can be estimated using each voxel that belongs to it.}
\vspace{-1.5em}
\label{fig:voxelize}
\end{figure}
}

\newcommand{\figresult}{
\begin{figure}[btp]
\centering
\includegraphics[width=0.85\linewidth]{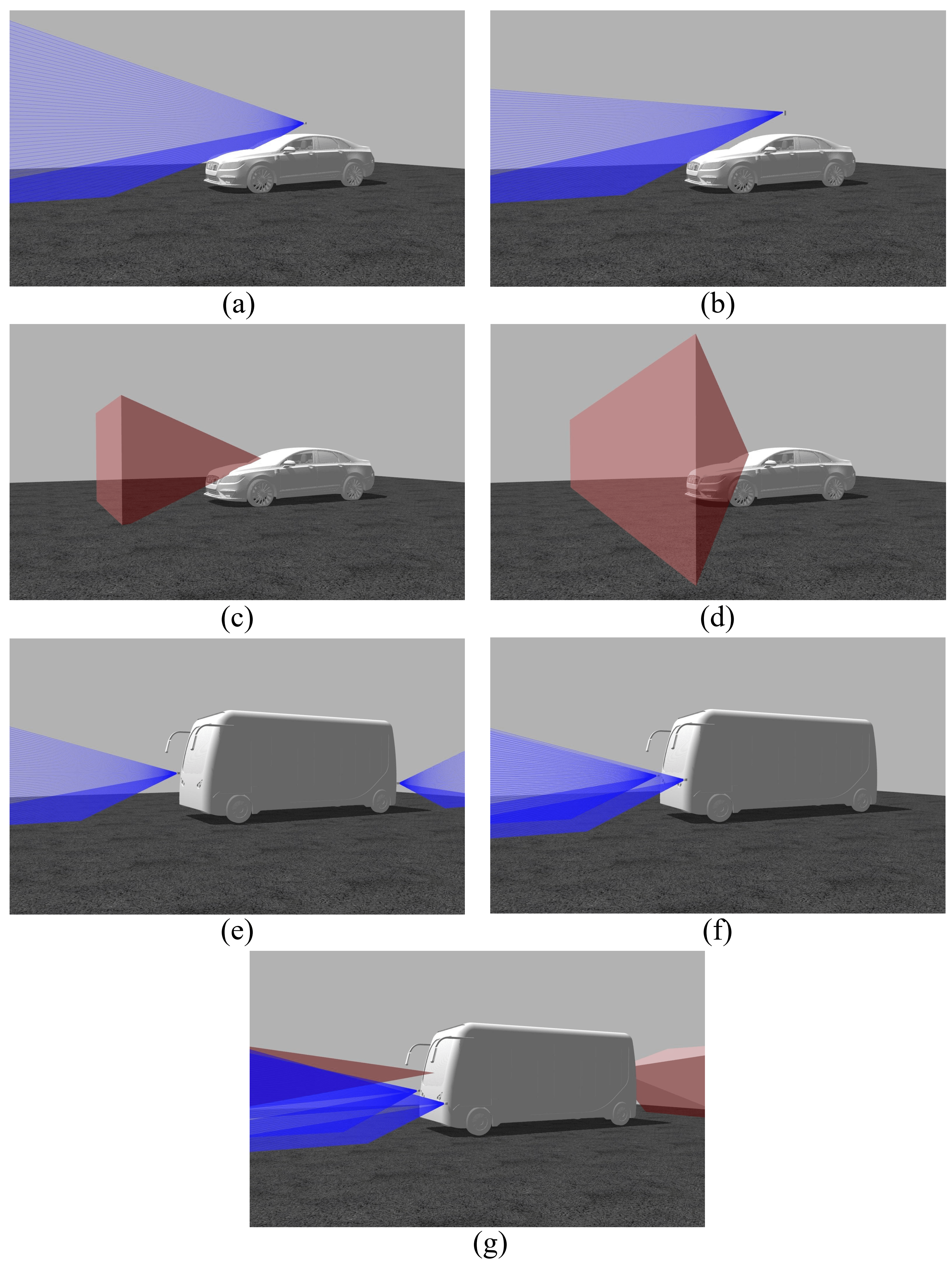} 
\vspace{-1em}
\caption{Simulation of the optimal sensor placement for different sensor selections. }
\vspace{-1.5em}
\label{fig:result}
\end{figure}
}

\newcommand{\figpitch}{
\begin{figure}[btp]
\centering
\includegraphics[width=0.85\linewidth]{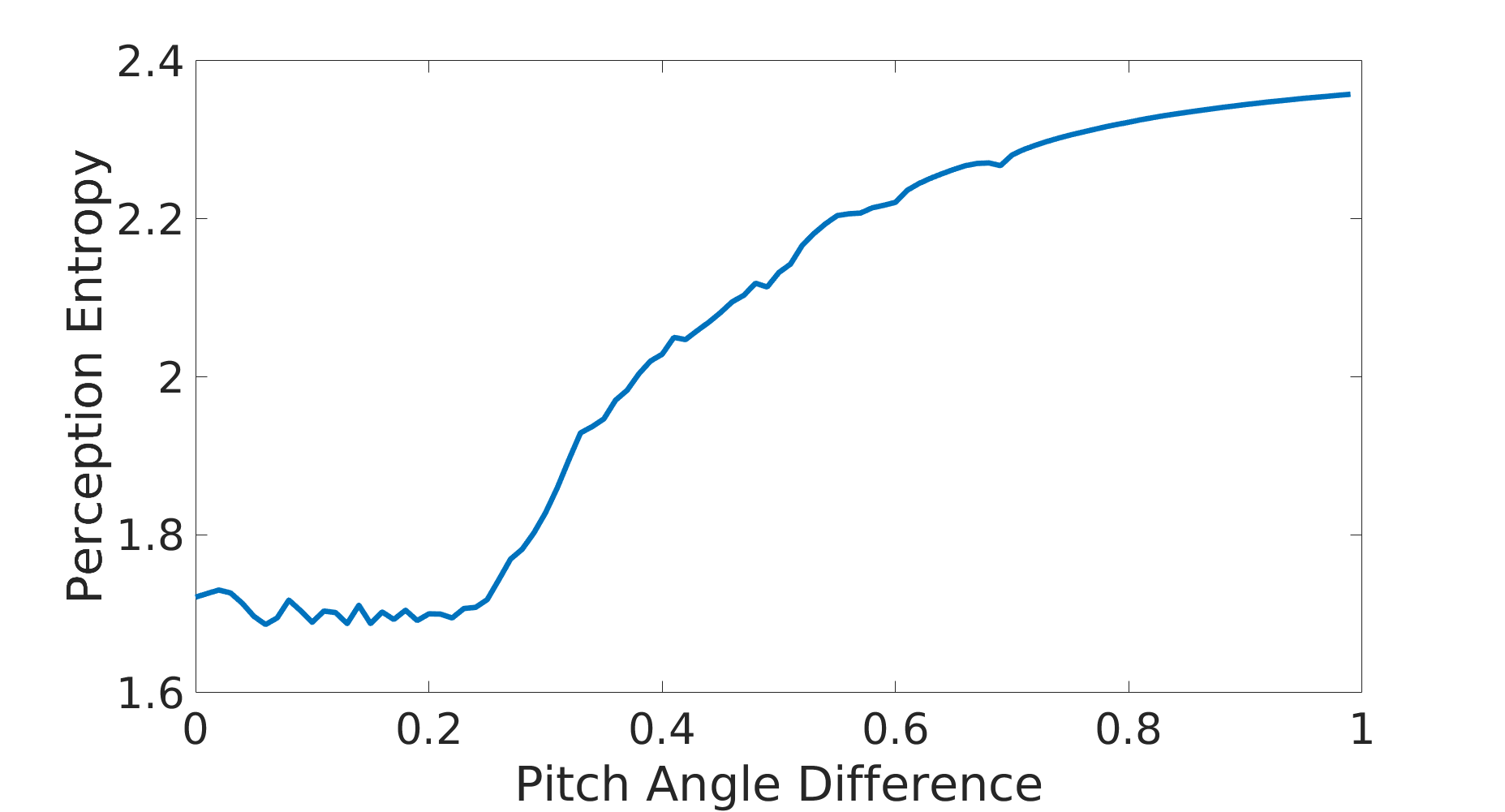} 
\vspace{-1em}
\caption{Relationship between the perception entropy and the pitch angle difference between the two LiDARs with the same mounting height. }
\label{fig:pitch}
\end{figure}
}

\newcommand{\etal}{et al.}

\title{\LARGE \bf
Perception Entropy: A Metric for Multiple Sensors Configuration Evaluation and Design
}

\author{Tao Ma$^*$, Zhizheng Liu$^*$, and Yikang Li$^{\dagger}$ 
\thanks{$^{*}$ Equally contributed to the work.}
\thanks{$^{\dagger}$ Corresponding author.}
\thanks{Tao Ma, Zhizheng Liu, and Yikang Li are with Autonomous Driving Group, SenseTime, Shanghai, China. {\tt\small \{matao, liuzhizheng, liyikang\}@senseauto.com}}
}

\begin{document}

\maketitle
\thispagestyle{empty}
\pagestyle{empty}

%%%%%%%%%%%%%%%%%%%%%%%%%%%%%%%%%%%%%%%%%%%%%%%%%%%%%%%%%%%%%%%%%%%%%%%%%%%%%%%%
\begin{abstract}
Sensor configuration, including the sensor selections and their installation locations, serves a crucial role in autonomous driving. A well-designed sensor configuration significantly improves the performance upper bound of the perception system. However, as leveraging multiple sensors is becoming the mainstream setting, existing methods mainly focusing on single-sensor configuration problems are hardly utilized in practice. 
To tackle these issues, we propose a novel method based on conditional entropy in Bayesian theory to evaluate the sensor configurations containing both cameras and LiDARs.
Correspondingly, an evaluation metric, perception entropy, is introduced to measure the difference between two configurations, which considers both the perception algorithm performance and the selections of the sensors. 
To the best of our knowledge, this is the first method to tackle the multi-sensor configuration problem for autonomous vehicles.
The simulation results, extensive comparisons, and analysis all demonstrate the superior performance of our proposed approach.
%Besides, highly efficient computation of the proposed method makes it practical for both academic or industrial use in autonomous driving area.

\end{abstract}

%%%%%%%%%%%%%%%%%%%%%%%%%%%%%%%%%%%%%%%%%%%%%%%%%%%%%%%%%%%%%%%%%%%%%%%%%%%%%%%%
\section{INTRODUCTION}
% \textcolor{red}{It's important to address our advantage of calculating and designing the single and whole sensor schemes.}
% \textcolor{red}{It's quite difficult to measure whether a sensor is located well or not with quantitative methods and metrics. And it's similar to evaluate a whole sensor scheme, the trade-off between perception capability and cost.
% How to define the upper bound of perception possibility of a specific sensor or a sensor scheme.}
% The detection algorithm doesn't decrease like the original input data, such as pixel area, beam and points numbers.
% \textcolor{red}{Multi sensor, different weather conditions, quantitative metrics.}

With the development of artificial intelligence, autonomous driving is attracting increasingly more attention in both industry and academia.
For autonomous vehicles~(AVs), the perception module, as the most upstream part, is vital for downstream tasks like localization, prediction, decision making, and path planning. 
Of the perception system, the sensor configuration plays the most crucial and fundamental role, which determines what can be perceived by the AVs, i.e. the upper-bound of the perceptions. Therefore, before developing an autonomous driving system, it is important to design a sensor configuration that can provide powerful perceptual abilities of the surrounding environments. 

%However, little interest is paid to the origin of data, sensors themselves, which play a vital role in the perception ability as they can provide varieties of rich information of the environment.
As sensing technologies grow rapidly, there exist various types of sensors that can be adopted~\cite{carballo2020libre}. Modern AVs usually fuse multiple sensors with different modalities, such as LiDAR, camera, and radar, to achieve robust perception performance for objects of different sizes under different conditions. How to select the set of the sensors and place them at the correct positions is a critical problem for the AV architect. 

%The most challenging part in sensor scheme design is quantitatively assessing its performance. %This performance is not based on the perception results. It only relies on the sensor setting itself. 
%It measures how much information around the vehicles that can be captured, which decides the ceiling of the perception system. 
%Differently, the perception algorithms are translating these raw information into the knowledge that can be leveraged by the system. 
%Therefore, decoupling the sensor scheme evaluation from the perception algorithm can help to shield the influence from the deep learning models and refine the sensor scheme design dedicatedly. 

% Take LiDARs as an example, Velodyne and Hesai both have types of different number of laser beams LiDAR such as 16-channel and 64-channel LiDAR, besides, the vertical angular resolution of Velodyne 64-channel LiDAR is 0.635\degree while that of Hesai 64-channel LiDAR is non-uniform (from 0.167\degree to 5\degree).
Few studies are investigating the sensor configuration evaluation and design. The mainstream solutions are heuristic approaches, which are based on the human experiences combined with vehicle information and the field of view~(FOV) of the sensors. 
%Therefore, the selection of sensors and following installation placement of these sensors is a challenging problem, but not studied extensively yet.
% However, to the best of our knowledge, few work has been done to address the problem of evaluating the perception performance of the full sensor scheme. 
They usually rely on straight-forward manual calculations, for example, adjusting the height of the LiDAR mounted on the top of vehicles to reduce the blind spot by the occluding roof~\cite{open_autonomous_driving_2019}. 
However, these kinds of solutions only roughly formulate the perception potential as a binary problem, i.e. whether the space within or beyond the FOV of the sensor. Even within the FOV, the number of pixels or laser points also significantly influences the results. Thus, the approaches are too rough to quantitatively evaluate the perception abilities of the sensor configuration. 
In addition, some works are focusing on the LiDAR configuration approaches~\cite{mou2018optimal, liu_2019}. However, these kinds of methods cannot be applied to other kinds of sensors.
As a result, we aim to formulate a unified framework to quantitatively evaluate the perceptivity of different sensor types and optimize the entire sensor configuration design based on the proposed evaluation metric. 
%As a result, the perception performance is only evaluated through running perception algorithms on the data collected from the real-world under a determinate sensor scheme. This gives great limitations to the design space of the sensor scheme and may lead to sub-optimal sensor solutions. 

% \textcolor{red}{This paragraph need to follow some fact that algorithm performance increase after changing the sensor scheme,}
% To completely utilize the potential of each sensor and optimize their joint perception ability upper bound, careful design choices of the sensor scheme should be made regarding the cost, the modality and the performance of sensors, as well as the installation placement of each sensor.

% Inspired by previous research on camera model \cite{zhang2000flexible}, we know that the filed of views (FOV), camera intrinsic parameters and activate pixels are correlated to each other, so the pixel area of objects is defined as a quantitative measurement for camera perception ability estimation. Similarly, we utilize beam numbers and point numbers as a statistic measurement for LiDAR perception ability.

In this paper, the perception potential of sensors is modeled as the theoretical possibility of objects being accurately perceived in the surrounding environment.
We address the sensor configuration optimization problem within the conditional entropy framework of Bayesian possibility theory and formulate the detected targets as a Gaussian distribution conditioned on sensor configurations.
Based on this formulation, a new metric, \emph{perception entropy}, is introduced to measure the performance of different sensor configurations. It is calculated by estimating the uncertainty of the detected targets under different configurations. The higher the entropy is, the lower the uncertainly is and, correspondingly, the better the specific configuration will be.
Therefore, our goal becomes finding the configuration with the highest perception entropy.
Additionally, our method covers sensors of different modalities elegantly by utilizing different reasonable sensor fusion strategies.
 
The main contributions of this paper could be summarized as four-fold: 
\textit{(i)} We propose a novel metric, perception entropy, to evaluate the perception potential of a particular sensor configuration, which can be used for sensor configuration design containing multiple sensors of different modalities.
\textit{(ii)} We find that the average precision~(AP) of the perception algorithms has a specific relationship with the number of pixels or cloud points on the targets. Then we use this feature to estimate the uncertainty of the targeted objects based on different sensor configurations.
\textit{(iii)} Reasonable sensor fusion strategies are proposed to unify different types of sensors into a single framework, which makes our method suitable for multi-modal sensor configuration design.
\textit{(iv)} To the best of our knowledge, this is the first method to tackle the multi-modal sensor configurations problem for AVs.
% \textit{(i)} We propose a conditional entropy based method to evaluate the perception ability of a particular sensor scheme, which can be used to design an optimal sensor scheme, including sensor specifications and installation placement. \textit{(ii)} The newly defined quantitative metric can reflect subtle differences among different sensors and sensor schemes. \textit{(iii)} Reasonable sensor fusion strategies are utilized to cover multiple sensors state so that the trade-off between perception potential and redundancy is well-considered. \textit{(iv)} To the best of our knowledge, this is the first method to tackle the multi-modal sensor configurations problem for AVs.
 
\section{RELATED WORK}

\subsection{Optimize Sensor Configuration}
Existing methods of optimizing sensor configurations only focus on sensors of a single modality.
\cite{Dybedal2017OptimalPO} found the optimal 3D sensors mounting positions to achieve the largest field of view through mixed-integer linear programming.
\cite{Rahimian2017OptimalCP} optimized the camera's pose for motion capture systems by computing target point visibility in the presence of dynamic occlusion from cameras.
To decrease the computation burden of previous mixed-integer programming methods, ~\cite{Dybedal2020GPUBasedOO} tried to solve the optimal placement of 3D sensors by exploiting the parallel processing and 3D architecture of a CUDA-based GPU.
In the field of autonomous driving, Mou \etal~\cite{mou2018optimal} firstly proposed a lattice-based approach combining with a cylinder-based approximation to solve the LiDAR configuration model. However, this method suffers from the curse of dimensionality and computation cost, which makes it unpractical for the real application.
Liu \etal ~\cite{liu_2019} extended this work and proposed a new bionic metric named volume to surface area ratio (VSR) to analyze the trade-offs between perception capability and design cost, but their modeling structure is still complex for real-world application and hardly reaches the global optimum despite simplifying the representation.
Besides, there seems little relationship between their bionic metric VSR and the performance of perception algorithms, which also severely restricts the deployment of this approach to practical AVs' design.

Researchers also studied optimal sensor placement in the general setting of solving a linear inverse problem, which aims at determining the model parameters that produce the sensor  measurement, mainly through greedy algorithms~\cite{ranieri2014near,jiang2016sensor,jiang2019group}. But for sensor placement on AVs, it's hard to represent the sensor  measurement and configurations as state vectors, which is not suitable for applying those algorithms. 

\subsection{Information Theory}
Approaches have also been proposed using standard Bayesian information theory to find the optimal sensor placement in other applications. Wang et al. modeled a sensor's potential of reducing the target state uncertainty using the entropy of the projection of the current state on the sensor subtracted by the entropy of the sensor noises for the task of target localization and tracking~\cite{wang2004entropy}. Their method only deals with the selection of one sensor in a 2D environment with only one object. Yang et al. extended this work by considering the placement of multiple heterogeneous sensors~\cite{yang2013optimal}. They developed an optimal strategy that maximizes the determinant of the Fisher information matrix by assuming the target has a Gaussian Prior. Furthermore, Krause et al. addressed the problem of optimal sensor placement in Gaussian processes of spatial phenomena by maximizing the mutual information between sensor locations~\cite{krause2008near}.   \cite{naeem2009cross} applied  cross-entropy optimization to the problem of selecting a subset of sensors to minimize parameter estimation error. These methods assume the perception results are deterministic by a fixed function. However, the perception algorithms used in AVs are usually based on deep neural networks, and hence the perception potential can be hardly optimized by those techniques. In this work, we propose a novel metric that incorporates the information theory, the detection performance of the neural networks, and sensor measurement together to formulate a unified framework for multiple sensors configuration design and evaluation.

\section{METHODOLOGY}
In this section, we first introduce the two concepts, \textit{perception space} and \textit{sensor measurement}, and review the basic knowledge of conditional entropy. 
Perception space defines the space we care about in the perception problem. Sensor measurement refers to the information of the target we can obtain for the sensor configuration, like the cloud points or pixels. 
Then we formulate the sensor configuration optimization problem within the conditional entropy framework. Under this formulation, perception entropy is introduced to measure the uncertainty of the perception results conditioned on the sensor configuration and corresponding sensor measurement of the targets. 
Thus, we can get the perception entropy for a single sensor. 
Finally, the fusion strategy is proposed to achieve a reasonable evaluation for a sensor configuration containing different types of sensors. 

\subsection{Perception Space}
To simplify the sensor configuration problem and simulation calculation, it is common to set up a perception space where the perception system is operated~\cite{mou2018optimal, liu_2019}. We only consider the perception potential within this space. Here we follow the conventional ground coordinate system $\mathrm{G}$ with the $x$-axis pointing forwards, the $y$-axis pointing leftwards, and the $z$-axis pointing upwards, with the origin lying at the center of the $x-y$ plane, where the AV is placed. As shown in Fig.~\ref{fig:space}, the AV is visualized in the perception space to determine exact sensor installation candidates, for more fine-grained simulation especially nearby the vehicle.

\figspace

\subsection{Sensor Measurement}
\label{subsec:meas}
We introduce a new concept, sensor measurement, to represent the perceptual signals from the targeted object to the candidate sensor configuration. It is an observation of the target for the specific setting, which depends on two factors, the target and the sensor configuration. 
For LiDAR, it denotes the 3D scanning point clouds on a targeted object with the location under LiDAR coordinate and intensity information of each reflected point.
% and we compute the number of laser points reflected from objects as the measurement intuitively.
For camera, it denotes a cluster of pixels that covers the target with color or darkness information.
% so the pixel area occupied by the object is treated as the measurement. 
% Although simulating the exact sensor  measurement would be more accurate, it greatly increases the complexity of the problem and requires very precise simulation of the environment. Therefore, we only consider part of the sensor  measurement.
%For autonomous driving scenarios, there are various types of objects to perceive such as cars, pedestrians, cyclists, \textit{etc}.
% Naturally, for each type of object in every location of this perception space, we can simulate the sensor measurement under the corresponding sensor configuration. 
%When calculating the sensor measurements of these objects, we regard different targets as cubes with different sizes for simplicity.
We notice that the main 3d detection algorithms for LiDAR and monocular camera usually use 3D bounding boxes outlining the points or pixels to define where is the candidate object. 
Empirically, compared to the fine-grained information like the intensity of the cloud points, the RGB of pixels, or their distributions, the sensor configuration mainly influences their numbers.
% In addition, wee notice that the main 3d detection algorithms for LiDAR and monocular camera usually use 3D bounding boxes outlining the points or pixels to define where is the candidate object. 
% In addition, the main 3D detection algorithms on LiDAR and camera usually use 3D bounding boxes outlining the candidate points or pixels as the  and output format.
% such as \cite{lang2019pointpillars,zhou2017voxelnet} and  usually
Therefore, we make the assumption\footnote{We mark all the assumptions used in our work with footnote.} that the sensor measurement only cares about the number of the pixels or cloud points.
In this paper, when using \emph{sensor measurement}, it is equivalent to the number of pixels or cloud points on the targeted objects.

% For LiDARs, we compute the number of laser points that would reflect from the voxel using $q$ and $x$. And for cameras, we compute the pixel size of the voxel in image.
% We denote these partial  measurement as g(z) and they can be calculated using basic geometry without a simulator, which greatly improves  efficiency.
%For autonomous driving scenarios, there are various types of objects to perceive such as cars, pedestrians, cyclists, \textit{etc}.
% Naturally, for each type of object in every location of this perception space, we can simulate the sensor measurement under the corresponding sensor configuration. 
%For simplicity, when calculating the sensor measurements of these objects, we regard them as cubes with different sizes. 
%Although u the entire exact sensor measurement would be more accurate and close to reality, it would be pretty laborious and computationally inefficient.
%Additionally, the ground-truth labels used in the detection algorithms are often in the form of a 2D/3D bounding box matching the pixels or points belonging to the targets.
%Therefore, we only consider part of the sensor  measurement and choose the most relevant factor which is affected by sensor configurations and  object locations: (1) the number of laser points reflected from the object and (2) the pixel area in the image occupied by the object.

The impact of sensor configuration on perception potential is reflected in the sensor measurement of a specific object. For example, if we adopt a camera with higher resolutions, it obtains more pixels in the captured image for the targeted object, which makes it more possible to detect the object. Similarly, compared to a 64-channel LiDAR, the fewer laser beams of a 16-channel LiDAR would produce more sparse and incomplete scanning results, especially at distant ranges.
Therefore, the configuration consists of two parts, sensor selections, and their mounting positions. The former relies on the type of the sensor, \textit{i.e.}, different laser beam numbers and maximum range of LiDAR, different FOV, and active resolution of the camera.
The latter can be characterized by a sensor's extrinsic parameter with respect to the ground coordinate system $\mathrm{G}$, which consists of the translation $\mathbf{t} = [t_x,t_y,t_z]^T$, and the rotation $\mathbf{R}$ represented by three Euler angles: $($\textit{roll, pitch, yaw}$)$. 

For the LiDAR measurement, we need to compute the formula of each beam to determine whether it goes through the target. Hence the zenith angle $\theta$ and the azimuth angle $\phi$ of each beam, as well as its maximum range, are required. Its direction $\mathbf{v}^{\mathrm{G}}$ in the ground coordinate system $\mathrm{G}$ can be computed as 
\begin{equation}
\label{eqn:1}
    \mathbf{v}^{\mathrm{G}} = \mathbf{R}^{-1}\cdot [\mathrm{sin}(\theta)\mathrm{cos}(\phi),\mathrm{sin}(\mathrm{\theta})\mathrm{sin}({\phi}),\mathrm{cos}({\theta})]^T .
\end{equation}
For the camera measurement, we need its intrinsic matrix $\mathbf{K}$ to project a target to the image, which could be approximately estimated through the horizontal FOV~(HFOV) $r_h$  of camera lens and its active resolution $H \times W$ as follows:
\begin{equation}
\label{eqn:2}
    \mathbf{K} = \begin{bmatrix}
\frac{W}{2tan(\frac{r_h}{2})} & 0 & \frac{W}{2}\\
0 & \frac{W}{2tan(\frac{r_h}{2})}  & \frac{H}{2} \\
0 & 0 & 1
\end{bmatrix}.
\end{equation}
For any point $\mathbf{p}^\mathrm{G}$ in the ground coordinate system $\mathrm{G}$, its projected pixel location $\mathbf{p}^\mathrm{C}$ can hence be obtained as:
\begin{equation}
\label{eqn:3}
    \mathbf{p}^\mathrm{C} \equiv \mathbf{K} \cdot (\mathbf{R} \cdot \mathbf{p}^\mathrm{G} + \mathbf{t}),
\end{equation}
and the pixel numbers of the target could be easily calculated subsequently.

Additionally, we simplify the problem by voxelizing each type of object into small cubes with a side length of $0.1$ m, then the perception potential of each object can be estimated by that of voxels belonging to the object for both LiDAR and camera as shown in Fig. \ref{fig:voxelize}.
For each voxel, its state is a three-dimensional vector $\mathbf{s} = [s_x,s_y,s_z]^T$ to represent its position in the ground coordinate system $\mathrm{G}$. Here we ignore the rotation part since it would barely affect sensor  measurement. 
\figvoxelize

% The impact of sensor configuration on perception potential is reflected in sensor  measurement intuitively. For example, if we place a camera closer to an object, the object would occupy more pixels in the captured image and thus resulting in a higher possibility being detected. Similarly, compared to a 64-channel LiDAR, the less laser beams of a 16-channel LiDAR would produces more discrete and uncompleted scanning results especially at distant ranges.

We denote  the configuration of a sensor as $q$, which includes both the sensor selection and its placement. The  measurement of a voxel is denoted as $m$, which is determined by the state of the voxel $\mathbf{s}$ and the sensor configuration $q$, as well as noise in the environment such as the light condition and dust. Here we assume\footnotemark{} the effect of noise on sensor  measurement $m$ is very small. Hence the noise is ignored in our calculation and the sensor measurement can be modeled as follows: 
\begin{equation}
    m = f(\mathbf{s}, q),
\end{equation}
where $f$ is a function that outputs the accurate sensor measurement when the object state and sensor configurations are given, using \eqref{eqn:1}\eqref{eqn:2}\eqref{eqn:3}.

%, which reduces the huge calculations for each cube from power level to linear. Additionally,  .... 
 
% Our method can be summarized as the following steps. First of all, the perception ability is interpreted as the perception of small cubes in the space whose prior distribution is obtained from the data set. Next we formulate the problem as the conditional entropy of the position of the cube given sensor  measurement with fixed location of sensors. After that, we estimate the posterior distribution of the cube using Gaussian distribution whose variance is approximated by sensor properties. Finally, we fuse distribution of different sensors and compute the final conditional entropy. 

\subsection{Conditional Entropy Theory and Formulation}
% We first review the concept of conditional entropy in the standard Bayesian probability theory.
The conditional entropy of a random variable $U$ on another random variable $V$ is a measure of the amount of uncertainty in $U$ if we have some knowledge about $V$. The conditional entropy is defined as follows using the probability distribution $p_V$ and the joint probability distribution $p_{(U,V)}$
\begin{equation} 
     H(U|V) = -\int_{\mathcal{V}} \int_{\mathcal{U}} p_{(U,V)}(u,v) \mathrm{ln}\left(\displaystyle\frac {p_{(U,V)}(u,v)}{p_V(v)}\right)\,du\, dv .
\end{equation}
We can further rewrite the formulation of the conditional entropy $H(U|V)$ as follows:
\begin{equation}
\label{eqn:CE}
    \begin{split}
     H(U|V) &= -\int_{\mathcal{V}} \int_{\mathcal{U}} p(u|v) \mathrm{ln}(p(u|v))\, du \;p(v)\,dv \\
           & = E_{v\sim p_{V}} H(U|v) .
    \end{split}
\end{equation}
Hence in this formulation, the conditional entropy can be interpreted as the expectation of the uncertainty of $U$ given $v$ with $v$ taken from the distribution $p_{V}$. 
In other words, the smaller the conditional entropy is, the more certain we are about $U$ after observing $V$.

\subsection{Perception Entropy}

Our method is based on a conditional entropy framework, by considering sensor measurement $M$ and the state of a voxel $\mathbf{S}$ as random variables.
First of all, $\mathbf{S}$ is defined as the position of a voxel from the pre-defined perception space whose prior distribution is obtained from the real data.
Next, given the target configurations $q$, the conditional entropy $H(\mathbf{S}|M,q)$ of the state of the voxel $\mathbf{S}$ on sensor  measurement $M$ is adopted as our metric, and we call it \textit{perception entropy}. The smaller the perception entropy is,
the more certain we are about the voxel position $\mathbf{S}$ after taking  measurement $M$, and the better perception potential the sensor configuration has. Following \eqref{eqn:CE}, we have: 
\begin{equation}
     H(\mathbf{S}|M,q) = E_{m\sim p_{M|q}} H(\mathbf{S}|m,q).
\end{equation} 
As $m$ is represented by a function $f(\mathbf{s},q)$ from our sensor measurement model, we can change the random variable of the expectation from $M$ to $\mathbf{S}$ using the law of unconscious statistician, so the conditional entropy becomes:
\begin{equation}
     H(\mathbf{S}|M,q) = E_{\mathbf{s}\sim p_\mathbf{S}} H(\mathbf{S}|m=f(\mathbf{s},q),q).
\end{equation} 

To compute this conditional entropy, the prior distribution of $p_\mathbf{S}$ is required, which describes the probability density that a voxel at position $\mathbf{S}$ is occupied by an object. $p_\mathbf{S}$ should depend on the actual distribution of objects in the environment. In the presence of a large dataset, this can be directly estimated by voxelizing each type of object in the dataset, and we denote the estimated distribution of $\mathbf{S}$ for an object type $c$ by $p_{\mathbf{S}_c}$.  

Moreover, the actual distribution of objects in the environment does not always meet the perception requirements. In some tourist attractions, perceiving objects in front of the vehicle is more important than the rear part as the vehicle is mostly driving forwards. Or the vehicle cares more about perceiving small objects like cones to perform obstacle avoidance.
Therefore, we add a weighting factor: $w(\mathbf{s},c)$, to represent the perception focus on different areas and types of objects. The final distribution $p_\mathbf{S}$ can be expressed as:
\begin{equation}
    p_\mathbf{S}(\mathbf{s}) = \eta \sum_c w(\mathbf{s},c) p_{\mathbf{S}_c}(\mathbf{s}),
\end{equation}
where $c$ is the specific object type, and $\eta$ is the normalizing factor.

Finally, we uniformly sample $\mathbf{s}$ within the perception space with an interval of $0.1$ m and compute the sensor  measurement $m = f(\mathbf{s},q)$  as well as the entropy of $\mathbf{S}$ given $m$ and $q$ at each position. The final perception entropy can thus be calculated with a weighted average of all entropy using distribution $p_\mathbf{S}$.  

\subsection{Gaussian approximation}
For simplicity, we model the probability distribution $p(\mathbf{S}|m,q)$ as a Gaussian distribution. The mean of this Gaussian distribution is unbiased and lies at the origin position $\mathbf{\tilde{s}}$ in distribution $p_\mathbf{S}$ with $\mathbf{\tilde{s}} = [\tilde{s}_x,\tilde{s}_y,\tilde{s}_z]^T$. As for the covariance, since for objects on the ground plane, their elevation is constant and for most perception algorithms the error of the perception results in the $z$ direction should be much smaller than that in the $x$ and $y$ direction, so we fix $\mathbf{S_z}$ as $\tilde{s}_z$. Furthermore, we treat the standard deviation in the $x$ and $y$ direction independently as the same standard deviation $\sigma$, so the distribution is symmetric. The resulting distribution is a 2D Gaussian distribution as follows:
\begin{equation}
    p((S_x,S_y)|m , q) = \mathcal{N}(\bm{\mu} = (s_x,s_y), \mathbf{\Sigma} = \sigma^2 \mathbf{I}).
\end{equation}
The entropy of this Gaussian distribution is calculated as 
\begin{equation}
\label{eqn:sigma}
    H(\mathbf{S}|m,q) = 2\ln(\sigma) +1+\ln(2\pi),
\end{equation}
which is only determined by the standard deviation $\sigma$.

Till this point, all our discussions are under the framework of probability theory, where entropy represents the difference in information and is only a \emph{relative} measure of the uncertainty. Therefore, the next step is to give the perception entropy a practical meaning that can be used to distinguish the performance between different sensor configurations. What's more, as it's meaningless to directly compare the number of laser points and the pixel size, we also need to unify measurements from multi-modality sensors together under the same metric. As LiDARs and cameras are both mainly used for 3D object detection, we can relate them together through their 3D detection performance, and use the performance to reflect their perception potential. Therefore, we adopt the most popular  evaluation metric for 3D detection performance: \emph{average precision~(AP)}, to bridge the gap between measurement and perception potential. Since $\sigma$ represents the uncertainty of the target estimations, it is tightly connected to the detection performance of that target.  Intuitively, the higher AP is, the smaller $\sigma$ should be. Specifically, when AP reaches its maximum 1, 
the uncertainty, $\sigma$, is close to its minimum. When AP equals 0, the algorithm has can never detect the target and hence $\sigma$ approximates the infinity. For simplicity, we assume\footnotemark{} $\sigma$ and  AP have the following relationship:
\begin{equation}
\label{eqn:i}
    \sigma = \frac{1}{\text{AP}}-1.
\end{equation}

% \textcolor{blue}{this paragraph is removed forward.}
% For different types of sensors, their measurement data also come in different forms. Here we mainly consider two types of sensors: LiDARs and cameras. LiDARs emit laser beams and output a point cloud with a distance measurement of each beam. Cameras passively take in visible lights and output an image with color information of each pixel. Although simulating the exact sensor  measurement would be more accurate, it greatly increases the complexity of the problem and requires very precise simulation of the environment. Therefore, we only consider part of the sensor  measurement. For LiDARs, we compute the number of laser points that would reflect from the voxel using $q$ and $x$. And for cameras, we compute the pixel size of the voxel in image. We denote these partial  measurement as g(z) and they can be calculated using basic geometry without a simulator, which greatly improves  efficiency. 

% \textcolor{red}{here is actually the prior used in our method and perform as a determinated factor}

Next, we investigate the connection between measurement $m$ and  AP. Naturally, $m$ has a positive correlation with  AP: the more laser points and the larger pixel size the voxel has, the better perception performance should be.
However, the exact relationship between  AP and $z$ is not straightforward and can be hardly captured by pure hypothesis. For example, if we double the laser point number of an object at a near range to get a more dense result, its AP may only increase by a small margin and would saturate at its upper limit 1. In terms of sensor configuration, that means the improvement in perception potential by using more and better sensors may not always meet its cost. Also, AP is dependent on the specific 3D detection algorithm, which should affect the perception potential as well. In this work, we assume\footnotemark{} that the sensor configuration design is bound to the algorithm used for 3D detection, and hence the relationship between AP and $z$ can be estimated by the performance of the specific perception algorithm.  We adopt PointPillars~\cite{lang2019pointpillars} for 3D LiDAR perception and RTM3D~\cite{RTM3D} for 3D monocular camera perception and evaluate AP on KITTI moderate dataset \cite{Geiger2012CVPR} for each type of object, and meanwhile, we compute the  measurement $z$ for each type of object given the sensor configuration used in KITTI. As the size of different types of objects varies, we further normalize $z$ by the ratio between the surface area of the object and the surface area of the voxel.
After that, we fit a curve between the measurement of the voxel $m$ and  AP as shown in  Fig.~\ref{fig:curve}.
 As the fitted curve indicates that  AP has a linear relationship with the natural logarithm of $m$ for both LiDAR and camera, we model the correlation between AP and $m$ as follows: 
\begin{equation}  
\label{eqn:ap}
\begin{aligned}
    % AP50 \approx (0.152 ln(g(z)) + 0.659), for LiDARs, \\
    % AP50 \approx (0.083 ln(g(z)) + 0.245 ), for cameras. 
    \text{AP} \approx a \ln (m) + b . 
\end{aligned}
\end{equation}
And we assume\footnotemark{} that once the linear relationship fits, it would not change a lot if the principle of algorithms doesn't change dramatically, and which is also universal to the same type of sensors. Then for other detection algorithms, the correlation could be established through solving the regression problem on the coefficients $(a, b)$. 

Combing \eqref{eqn:sigma}\eqref{eqn:i}\eqref{eqn:ap}, we can finally compute $H(S|m,q)$ for a single sensor as:
\begin{equation} 
    H(\mathbf{S}|m,q) = 2\ln \left(\frac{1}{a\ln(m) + b} -1\right) +1+\ln(2\pi).
\end{equation}
In practice,  we clamp AP to be in the range of $[0.001,0.999]$ for numerical stability.  

\figcurve

\subsection{Sensor Fusion}
The perception systems of AVs usually contain both cameras and LiDARs. To fuse the  measurement of different sensors, state-of-the-art perception systems mainly adopt two strategies: \emph{early fusion} and \emph{late fusion}. 
The former mixed data from all sensors together and could be regarded as a \textit{Super Sensor}, while the latter usually operate on the layer of each sensor's independent perception result. The application of different fusion strategies depends on the specific type of sensors and the principle of perception algorithms. For different LiDARs, early fusion is preferred as point clouds could be combined directly as a merged point cloud. For fusion between LiDARs and cameras or different cameras,  late fusion is more reasonable, as the point cloud and the image are in different forms and images cannot be merged.

For the early fusion strategy, suppose we have $n$ sensors in total, and their configurations are denoted as $\mathbf{q} = (q_1,q_2,...,q_n)$. Then their  measurement $\mathbf{m} = (m_1,m_2,...,m_n)$, with $m_i = f(\mathbf{s},q_i)$. When the early fusion strategy is adopted, the sensor  measurement $m_i$ are accumulated:
\begin{equation}
    m_{fused} = \sum_i m_i,
\end{equation}
\textit{i.e.}, the numbers of laser points from different LiDARs are summed together. And the distribution of $\mathbf{S}$ is calculated using the accumulated  measurement $m_{fused}$.

For the late fusion strategy,
% \textcolor{yellow}{what's the meaning of this sentence?} we follow a heuristic basic principle: suppose there are two cameras detecting the vehicles in the same scenario, the one is 50\% and the other is 70\%, so the final possibility is $1-(1-0.5)*(1-0.7)=0.85$. 
we calculate the product of distributions of all sensors and normalize it to a new Gaussian distribution. Suppose the standard deviation of the normal distribution of the n sensors are $\mathbf{\sigma} = (\sigma_1,\sigma_2,...,\sigma_n)$,
then the product of these weighted normal distributions follows another Gaussian distribution after normalization, with standard deviation
\begin{equation}
\small
    \sigma_{fused} =  \sqrt{\displaystyle\frac{1}{\sum_i \frac{1}{\sigma_i^2}}}.
\end{equation}
 
In practice, there are multiple sensors with different modalities and these two fusion strategies are used simultaneously in the perception algorithm. Therefore, to incorporate both strategies in our calculation, we first sum the sensor  measurement for the part of early fusion and formulate each early-fused distribution, and then we compute the final entropy using all distributions for the late fusion part. Our fusion strategies in calculation make reasonable differences for sensor configurations and the details could be found in Sec. \ref{sec:experiments}.

Finally, we utilize the whole process to calculate the perception entropy and iteratively search for the optimal configurations with very high efficiency. For each sensor selection, its sensor placements are optimized using  Algorithm~\ref{alg:1} by random search. And the optimal sensor configuration is obtained by comparing and selecting the best sensor selection.

\alg

\section{EXPERIMENTS}
\label{sec:experiments}

The perception spatial space is set as a rectangular region, with $x \in [-80\text{ m}, 80\text{ m}]$, $y \in [-40\text{ m}, 40\text{ m}]$, and $z \in [0\text{ m}, 5\text{ m}]$. We consider five types of objects to perceive: cars, pedestrians, cyclists, trucks, and cones. For the prior distribution $p_\mathbf{S}$, we collect the per-class distribution $p_{\mathbf{S}_c}$ from our road test and the perception weights $w(c,\mathbf{s})$ is set to be 1 for all objects in all areas by default and will be adjusted for comparison. In our experiment, we also adopt PointPillars~\cite{lang2019pointpillars} for 3D LiDAR perception and RTM3D~\cite{RTM3D} for 3D monocular camera perception and the coefficients $(a,b)$ for \eqref{eqn:ap} are $(0.152,0.659)$ and $(0.055,0.155)$ respectively for those two algorithms.
The specifications of different LiDARs used in our experiments are summarized in Table~\ref{table:spec}. 
For cameras, the active resolution is fixed to be $1920\times 1080$ and we will only adjust the HFOV.
In the following experiments, we only search for $t_x,t_y,t_z$, and the \textit{pitch} angle of sensor placement, while the \textit{roll} and \textit{yaw} angle are fixed to be $0^{\circ}$ for symmetry of the placement.

Our implementation code is in C++ and all experiments are conducted on an Intel Core i7-8700 desktop CPU. The initial  neighborhood $N$ is set to be $[-1\text{ m},1\text{ m}]$ for translation and $[-30^\circ,30^\circ]$ for rotation. The final neighborhood $N_0$ is set to be $[-0.01\text{ m},0.01\text{ m}]$ for translation and $[-0.3^\circ,0.3^\circ]$ for rotation. For each neighborhood, we randomly generate $1000$ sensor placement, and the decay factor $\mathrm{k}$ is set to be $\frac{1}{2}$. The running time of our algorithm is less than a second for a specific configuration and the whole optimization takes around an hour, which is much faster than previous methods~\cite{mou2018optimal, liu_2019}.

\tablelidar

\figresult

\subsection{Velodyne HDL-64E v.s. Hesai Pandar64}
Here we consider two types of 64-channel LiDAR installing on a car.
The optimal installation placement for both LiDARs is listed in experiment \#1 of Table~\ref{table:result} and visualized in Fig.~\ref{fig:result}(a) and (b). The perception entropy of Pandar64 is $1.6429$, which is much smaller than HDL-64E, $2.1212$.
Compared to HDL-64E, we noticed that Pandar64 has a non-uniform vertical resolution, which is $0.167\degree$ in the middle channels.
Although the bottom channel laser of Pandar64 is blocked by the car hood,  this selection could result in more laser beams and points on targets at a farther distance, while the same result of testing is shown in~\cite{carballo2020libre}.
That could further prove the perception entropy could accurately reflect the minimal difference between sensors.
%Although Pandar64 has slightly worse horizontal resolution, even the bottom channel is blocked by the car hood......
% The HDL-64 LiDAR has uniform vertical resolution. However, the Pandar64 LiDAR has non-uniform vertical resolution, the vertical resolution is very large on both ends, while the vertical resolution is much smaller than HDL-64 in the middle channels(from -$6^{\circ}$ to $2^{\circ}$), which is specially designed for autonomous vehicles.
% As shown in experiment \#1 of table \ref{table:result}, although Pandar64 even has slightly worse horizontal resolution, its perception entropy is much smaller than HDL-64, which proves the effectiveness of its non-uniform horizontal resolution design. As shown in figure ?? (a), the optimal placement of Pandar64 even has intersect with the car hood, thanks to most of its beam are concentrated at the center.

\subsection{60\degree HFOV camera v.s. 120\degree HFOV camera}
% Here we consider installing two types of cameras on a car.
A camera with $60^{\circ}$ HFOV and another one with $120^{\circ}$ HFOV are compared in this experiment. We also double the weight $w(\mathbf{s},c)$ for $|s_x|>40\text{ m}$ to focus more on perceiving distant objects. The optimal installation placement for both cameras is shown in experiment \#2 of Table~\ref{table:result} and simulated in Fig. \ref{fig:result}(c) and (d).
The result shows the perception entropy of the 60\degree HFOV camera ($2.0055$) outperforms that of the 120\degree HFOV camera ($2.0237$). Intuitively, the smaller FOV camera has a larger focal length, resulting in a higher resolution of objects which is crucial for perception from a long distance but sacrifices a lot of horizontal views at a near distance.
That also validates that the perception entropy could
help us balance the perception potential of sensors for \textit{seeing far} and \textit{seeing wide}.
% accurately reflect the focus on difference areas in the perception space.

\subsection{Early Fusion of LiDARS}
\label{subsec:ef}
One LiDAR mounted on the top roof is unrealistic for an autonomous bus, so we explore the placement of multiple LiDARs and two Pandar40 LiDARs are chosen for the experiment.
% In this experiment, we install two Pandar40 LiDARs on an autonomous bus and the results using early fusion of LiDARS are shown in experiment \#3 of Table~\ref{table:result}.
In experiment \#3 of Table~\ref{table:result}, the first set is under the default setting, while the weight $w(\mathbf{s},c)$ for $s>0$ is doubled in the second set to focus more on the front area, and the optimal placements of simulation result are shown in Fig.~\ref{fig:result}(e) and (f) respectively.
% In the first part we keep the default setting while in the second part we double the weight $w(\mathbf{x},c)$ for $x>0$  to focus more on the front area.
The former has two LiDARs mounted on the diagonal line of the bus, which indicates that covering the whole 360\degree surrounding environments completely is the main factor for LiDAR configuration.
After we increase the weight of the front area, the second set has two LiDARs mounted on the front A-pillar of the bus.
% we can see that due to the change in the perception weight of front area, the optimal placement also changes completely.
% The first one has two LiDARs placed on the diagonal line of the bus, while the second places both LiDARs in the front.
Note that the optimal placement is not exactly symmetric to the bus center, and there is a small \textit{pitch} angle offset between these two LiDARs.
Intuitively, this could not only prevent the overlap of laser beams from different LiDARs, but one LiDAR's discrete space between two adjacent beams will be also inserted by the other's laser beam, contributing more laser points on the targets. This characteristic could be captured by our perception entropy shown in Fig. \ref{fig:pitch}.

\figpitch

\subsection{LiDAR-Camera Late Fusion}
In the final part, we consider real-world settings where multiple sensors of different modalities are used.  As shown in Fig. \ref{fig:result}(g), we extend the second configuration of Sec.~\ref{subsec:ef} with two cameras. The front camera has 30\degree HFOV to benefit the perception at far ranges, and the rear camera with 60\degree HFOV could perceive the areas that both LiDARs can't reach. And the perception entropy is much better than the LiDAR only configuration.

% the first setting uses two pandar40 LiDARs and a 60FOV camera. The second setting consists of  two pandar64 LiDARs and two 60 FOV cameras. We want to prove that the use of  more and better sensors may not meet its cost.

%For two LiDARs, we find that a minimal pitch degree offset between them will conduct the most informative perception. Figure X shows the extreme point on the measurement curve.
%Intuitively, one LiDAR's discrete space between two neighboured beams will be filled with the other's laser beam, which will produce a more dense scanning result of objects.

%\textcolor{red}{In this section, we mainly conduct the comparison as follows:a. Velodyne HDL-64E \textit{v.s.} Hesai Pandar64, the key point is the uniform vertical resolution.b. 2 Pandar40 \textit{v.s.} 1 Pandar 64, add the pitch experiment here.c. the difference between different FOV cameras.d. the fusion of lidar and camera.}

\tableresult
 
\section{CONCLUSIONS}
In this paper, we propose a metric named perception entropy to calculate the perception potential of a sensor configuration and iteratively optimize the installation configurations of sensors with very high efficiency. Perception entropy could accurately reflect the difference between any two sensors or sensor configurations,
% even  which will help us select the sensors quantitatively evaluate and compare
and provide the theoretical quantitative upper bound of a specific sensor configuration's perception potential.
Due to the thoughtful sensor fusion strategies, our method could cover multiple sensors elegantly, so that the trade-off between perception ability and redundancy could be well-considered and reached the most cost-effective.
Finally, the simulation results, extensive comparison, and analysis could demonstrate the superior performance of our proposed approach.

As the perception entropy in our calculation is related to the actual perception algorithms, if 
the principle of perception algorithms changes fundamentally, the definition of sensor measurement may change as well as its relationship with the algorithms' performance.
In the future, the perception entropy could incorporate the impact of environmental conditions, e.g., rains and fogs would weaken the signal of LiDAR and produce much more noise.
In addition, we could model the motion of different types of objects to better simulate the perception potential by introducing the concepts of tracking algorithms.
% When the performance of the perception algorithms changes, the optimal sensor configurations may change time with a minor variety simultaneously. Once the principle of algorithms changes a lot, ...

\bibliographystyle{IEEEtran}
\bibliography{egbib}

\end{document}